\documentclass[runningheads]{llncs}
\usepackage[T1]{fontenc}

\usepackage[dvipsnames]{xcolor}
\usepackage{amsmath}
\usepackage{amssymb}
\usepackage{adjustbox}
\usepackage{dsfont}
\usepackage{multirow}
\usepackage{tikz}
\usetikzlibrary{positioning}

\usepackage{graphicx}
\usepackage{hyperref}
\usepackage{color}

\begin{document}
%
\title{SODAWideNet++: Combining Attention and Convolutions for Salient Object Detection}
\titlerunning{SODAWideNet++}

\author{Rohit Venkata Sai Dulam \and Chandra Kambhamettu}

\authorrunning{RVS Dulam, C Kambhamettu}

\institute{Video/Image Modeling and Synthesis (VIMS) Lab \\ University of Delaware, Newark DE 19713, USA \\
\email{\{rdulam, chandrak\}@udel.edu}}

\maketitle              
\begin{abstract}

Salient Object Detection (SOD) has traditionally relied on feature refinement modules that utilize the features of an ImageNet pre-trained backbone. However, this approach limits the possibility of pre-training the entire network because of the distinct nature of SOD and image classification. Additionally, the architecture of these backbones originally built for Image classification is sub-optimal for a dense prediction task like SOD. To address these issues, we propose a novel encoder-decoder-style neural network called \textbf{SODAWideNet++} that is designed explicitly for SOD. Inspired by the vision transformers' ability to attain a global receptive field from the initial stages, we introduce the Attention Guided Long Range Feature Extraction (AGLRFE) module, which combines large dilated convolutions and self-attention. Specifically, we use attention features to guide long-range information extracted by multiple dilated convolutions, thus taking advantage of the inductive biases of a convolution operation and the input dependency brought by self-attention. In contrast to the current paradigm of ImageNet pre-training, we modify 118K annotated images from the COCO semantic segmentation dataset by binarizing the annotations to pre-train the proposed model end-to-end. Further, we supervise the background predictions along with the foreground to push our model to generate accurate saliency predictions. SODAWideNet++ performs competitively on five different datasets while only containing 35\% of the trainable parameters compared to the state-of-the-art models. The code and pre-computed saliency maps are provided at \url{https://github.com/VimsLab/SODAWideNetPlusPlus}.

\end{abstract}

\section{Introduction}
\label{sec:intro}
Salient Object Detection (SOD) requires identifying the objects that catch a viewer's immediate attention from visual data. Saliency is vital for many areas of Computer Vision, including Semantic Segmentation\cite{sun2022inferring}, Person Identification~\cite{Kim_2021_CVPRpersonsearch}, etc. The first methods for SOD \cite{cheng2014globalregionbased,jiang2013saliencyregionbased2} relied on hand-crafted priors like color, contrast, etc. With the advent of Deep Learning, the emphasis has shifted towards developing Deep Learning based solutions for SOD. 

The fundamental idea of Deep Learning based solutions for SOD involves building novel feature refinement modules that rely on semantic features extracted using ImageNet pre-trained backbones. However, these backbones are primarily designed for image classification and do not fully meet the intricate needs of SOD. SOD often involves analyzing images with multiple objects, unlike the single-object focus typical in image classification datasets. This discrepancy can result in less optimal saliency predictions. Additionally, the standard approach of only integrating the backbone with the refinement modules during the fine-tuning phase misses a critical chance to pre-train these components together, which could enhance overall model performance. To better address these challenges, our models are designed specifically for SOD from the ground up, allowing for the entire network to be pre-trained simultaneously. A crucial part of our approach involves adapting the COCO semantic segmentation dataset \cite{lin2014microsoftcoco} for SOD by converting the segmentation labels to saliency labels. Although significantly smaller than other pre-training datasets like ImageNet, our results illustrate the advantages of pre-training the entire model.

Moving toward the architectural innovations that underpin our model, it is crucial to recognize the dramatic shift from traditional Convolutional Neural Networks \cite{he2016deepresnet,simonyan2014veryvgg16} to Vision Transformers (ViTs) \cite{dosovitskiy2020vit}, which have significantly advanced the benchmarks across various computer vision tasks. Unlike CNNs, which are constrained by their local receptive fields and often fail to adapt to the unique characteristics of different inputs, ViTs excel by capturing global relationships and input-specific details through self-attention. As a consequence of the local receptive field, CNNs typically employ a hierarchical approach to extracting global features that rely on extreme downsampling of the input. While broadening the receptive field, this process results in a substantial loss of detail, a critical drawback for tasks requiring high-resolution outputs. Additionally, the convolution operation in CNNs is designed to detect common patterns across different instances, which does not suffice for capturing attributes unique to individual instances. Several studies \cite{dai2017deformable,zhu2019deformablev2} have attempted to overcome these limitations but often incur high computational costs. To furnish a CNN with the ability to utilize instance-specific information, we employ self-attention and use it to guide the features extracted by convolutional operations.

Pre-trained on the modified COCO dataset, we propose \textbf{SODAWideNet}++, a modified SODAWideNet \cite{dulam2023sodawidenet} architecture that seamlessly integrates Attention into convolutional components to extract local and global features. To extract global features, we combine dilated convolutions from the Multi-Receptive Field Feature Aggregation Module (MRFFAM) and Attention from Multi-scale Attention (MSA) modules and propose Attention guided Long Range Feature Extraction (AGLRFE). Furthermore, we modified the Local Processing Module (LPM) by adding Attention and thus proposed an Attention-enhanced Local Processing Module (ALPM) to extract local features. Finally, we reuse the Cross Feature Module (CFM) to combine features from both the proposed components. We summarize our contributions below - 

\begin{enumerate}
    \item We propose SODAWideNet++, a convolutional model utilizing self-attention to extract global features from the beginning of the network.
    \item We modify the famous COCO semantic segmentation dataset to generate binary labels and use it to pre-train the proposed model.
    \item We propose AGLRFE to extract features from large receptive fields using dilated convolutions and Self-Attention.
    \item Unlike prior works, we supervise foreground and background predictions for increased accuracy.
\end{enumerate}

\section{Related Works}

\subsection{Salient Object Detection with ImageNet pre-training}

PiCAN~\cite{liu2018picanet} developed a contextual attention module to attend to essential context locations for every pixel from Resnet-50 features. BASN \cite{qin2019basnet} uses an encoder-decoder-style network initialized by a Resnet-34 model and a boundary refinement network on top to produce accurate saliency predictions with crisp boundaries. (F3-N) \cite{wei2020f3net} uses a Resnet-50 network to extract semantic features refined by a cascaded feedback decoder (CFD) and cross-feature module to produce saliency outputs. VST \cite{Liu_2021_ICCV_VST} is the first work to propose a vision transformer-based SOD model. PSG uses a loss function that creates auxiliary saliency maps based on the morphological closing operation to generate accurate saliency maps incrementally. ET \cite{jing_ebm_sod21generative} uses an energy-based prior for salient object detection. RCSB \cite{Ke_2022_WACV_RCSB} refines features extracted from a Resnet-50 \cite{he2016deepresnet} backbone using Stage-wise Feature Extraction (SFE) and a few novel loss functions for SOD. CSF-R2 \cite{gao2020sod100kcsf} proposes a flexible convolutional module named gOctConv to utilize multi-scale features for SOD. EDNet \cite{wu2022edn} presents a novel downsampling technique to learn a global view of the whole image to generate high-level features for SOD. PGN uses a combination of Resnet and Swin\cite{liu2021swin} models to generate saliency maps. ICON \cite{zhuge2021salienticon} introduces a diverse feature aggregation (DFA) component to aggregate features with various receptive fields and increase the feature diversity. TR \cite{lee2022tracer} uses an EfficientNet backbone and attention-guided tracing modules to detect salient objects. LDF \cite{ldf} proposes a label decoupling framework to detect salient objects. The authors disintegrate the original saliency map into body and detail maps to concentrate on the central object and object edges separately. PA-KRN \cite{pakrn} uses a knowledge review network to first identify the salient object and then segment it. RMF \cite{rmformer} proposes a novel Recurrent Multi-scale transformer that utilizes a transformer and multi-scale refinement architectures for SOD. SR \mbox{\cite{SelfReformer}} proposes a novel framework that enhances global context modeling and detail preservation to generate accurate saliency predictions. VSC \mbox{\cite{luo2024vscode}} proposes a foundational model for SOD that uses programmable prompts to generate saliency predictions.

\section{Proposed Method}

\begin{figure*}
    \centering
    \includegraphics[width=\textwidth]{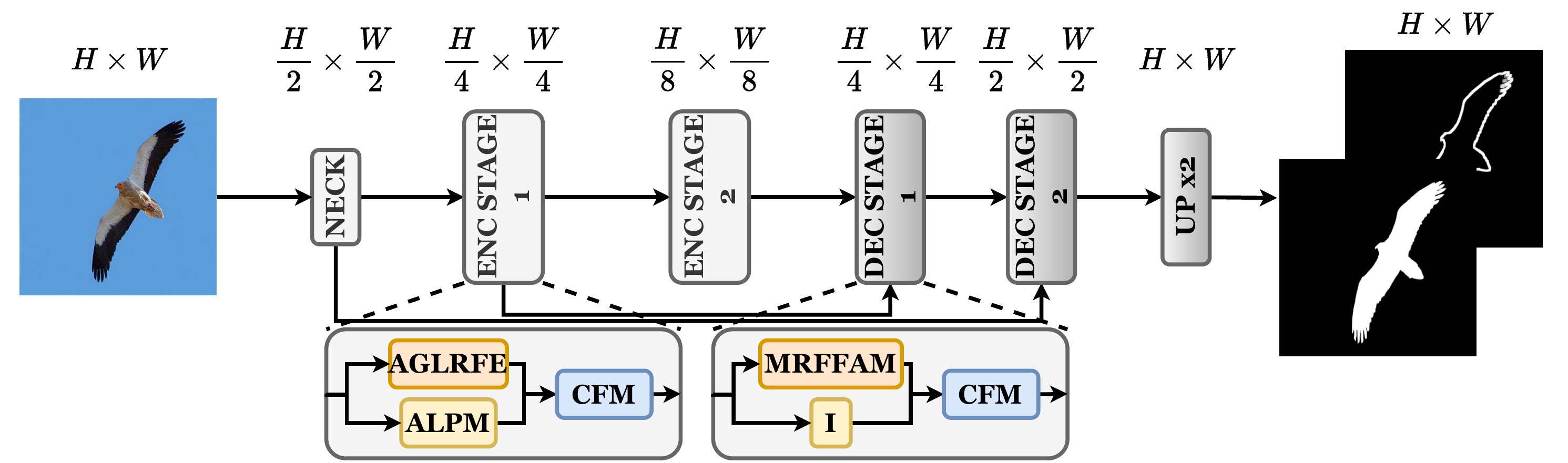}
    \caption{The proposed architecture \textbf{SODAWideNet++} contains two branches, one to extract global features using AGLRFE and the other to extract local features using ALPM. These global and local features pass through CFM, producing the output of an encoding layer. The decoding layers also consist of two parallel paths, MRFFAM, to decode features through multiple receptive fields and an Identity operation.}
    \label{fig:architecture}
\end{figure*}

The proposed model is an improved version of the design principles introduced in \cite{dulam2023sodawidenet}. This section will first explain the SODAWideNet \cite{dulam2023sodawidenet} model, which serves as the basis for SODAWideNet++. Then, we will introduce the core components of SODAWideNet++, namely the Attention Guided Long Range Feature Extraction (AGLRFE) and the Attention Local Pooling Module (ALPM). Finally, we will discuss the specific loss function used in our model and other essential elements of our design strategy.

\subsection{SODAWideNet}
SODAWideNet, as described in \cite{dulam2023sodawidenet}, employs an encoder-decoder architecture to generate saliency predictions. The encoder is composed of three parallel pathways: the Multi-Receptive Field Feature Aggregation Module (MRFFAM), the Multi-Scale Attention (MSA), and the Local Processing Module (LPM). These pathways are engineered to capture both global and local features concurrently. The MRFFAM extracts and consolidates semantic information from multiple receptive fields using large dilated convolutions, thereby strengthening the model's capability to identify objects of varying sizes. Similarly, MSA utilizes self-attention to extract global features in a hierarchical manner, keeping in mind the computational complexity of the attention operation. The LPM extracts local features using smaller $3 \times 3$ convolutions and multiple maxpooling operations. The Cross-Feature Module (CFM) merges MRFFA, MSA, and LP module features. This module contains a series of convolutions that selectively combine features due to the distinct nature of features obtained from the three components. Each convolution consists of a $3 \times 3$ convolution operation followed by a Group Normalization layer and GELU activation function. In the decoding phase, the architecture features two parallel paths, MRFFAM and Identity, which work together to decode the encoded features and generate the final saliency outputs. The overall architecture consists of two such encoder and decoder blocks. We reuse these decoder blocks in the proposed SODAWideNet++ model.

\begin{figure*}[h!]
    \centering
    \includegraphics[width=0.65\linewidth, height=7.5cm]{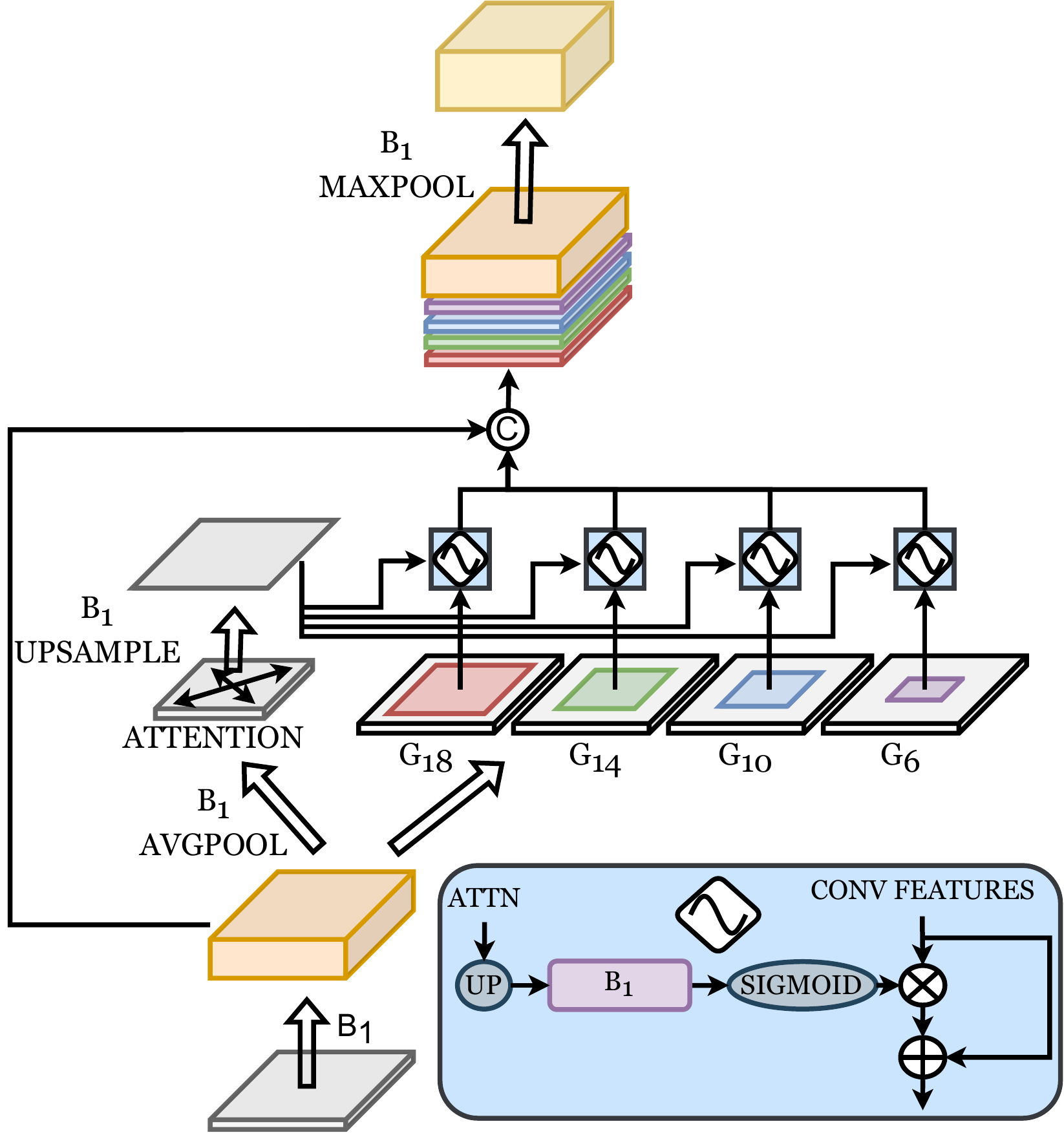}
    \caption{\textbf{Attention-guided Long-Range Feature Extraction module (AGLRFE)} consists of two branches: a collection of dilated convolutions with different dilation rates to extract long-range convolution features and a self-attention block. We reduce the spatial resolution of the input before the Self-Attention block using Average Pooling and refine them using a series of convolution operations. Attention features are then upsampled to the exact resolution as the convolution features from the dilated convolutions. Then, using a series of convolution layers, we bring its channel size to one and pass it through a Sigmoid layer. These features refine our long-range convolution features, thus inducing input-reliance.}
    \label{fig:aglrfe}
\end{figure*}

\subsection{SODAWideNet++}
SODAWideNet++ builds upon the foundational architecture of SODAWideNet, maintaining distinct branches for extracting local and global features. While SODAWideNet has demonstrated commendable performance when trained from scratch, it still lags behind state-of-the-art models in terms of overall efficacy. In SODAWideNet++, we refine the approach of capturing long-range features by merging the functionalities of dilated convolutions and attention into a single, streamlined module named Attention-Guided Long Range Feature Extraction (AGLRFE). This module is engineered to generate long-range, input-dependent convolutional features, thereby enhancing the robustness of the feature representation. Additionally, we incorporate attention into the Local Processing Module (LPM) to increase input dependency, rebranding it as the Attention-enhanced LPM (ALPM). The Cross-Feature Module (CFM) remains unchanged and is tasked with the integration and refinement of both local and global features. The decoder structures in SODAWideNet++ and SODAWideNet are identical, preserving the architectural consistency across both models.

\subsection{Attention guided Long Range Feature Extraction (AGLRFE)}
Vision transformers have achieved significant success due to the effectiveness of Self-Attention in extracting important semantic features across large receptive fields. On the other hand, conventional CNNs achieve a global receptive field by downsampling the input hierarchically, which can lead to the loss of critical features and increase the model's parameters. 

To overcome these limitations, MRFFAM employs multiple dilated convolutions with large dilation rates to expand the receptive field of our network. The input is divided into chunks, and each chunk is input to a dilated convolution with a specific dilation rate. The output of the MRFFAM block is to concatenate the outputs of dilated convolutions along with the input in the channel dimension. This resultant feature map is either sent through a downsampling layer or a series of convolutions, depending on whether it is in the encoder or the decoder.

However, the intrinsically input-dependent nature of Self-Attention is absent in a convolution operation, thus restricting the effectiveness of a CNN. Thus, to leverage the strengths of convolutions and Self-Attention, we propose a hybrid module, modifying the MRFFAM block to contain an Attention operation. Specifically, we use dilated convolutions to capture long-range convolutional features and Self-Attention to derive input-specific features and use them to guide the convolution features. Below, we illustrate particular components of AGLRFE followed by a sequence of operations.

\begin{gather*}
    B_{i}(x) = convB_{i}(convB_{i}(x)) \\
    G_{i}(x) = conv_{i}(conv_{i}(x)) 
\end{gather*}

$convB_{i}$ implies a $3 \times 3$ convolution with dilation $i$ followed by Batch Normalization and GELU activation function, whereas $conv_{i}$ implies a $3 \times 3$ convolution with dilation $i$ followed by Group Normalization and GELU activation function.

As seen in figure \ref{fig:aglrfe}, the input to an AGLRFE block \textit{X} is transformed using a series of $3 \times 3$ convolution layers. 

\begin{equation*}
    feat = B_{1}(X)
\end{equation*}

Next, there are two parallel paths: one path computes attention features, and the other computes convolutional features of different receptive fields. To compute the attention features, \textit{feat} is downsampled using an Average Pooling layer (\textit{AvgPool}), which is then transformed by a series of convolution layers.

\begin{equation*}
    f_{attention} = Attn(B_{1}(AvgPool(feat)))
\end{equation*}


\begin{equation*}
    Attn = softmax \left( \frac{K \times Q}{\sqrt{d_{k}}} \right) V
\end{equation*}

where $K, Q$ and $V$ are computed using $conv_{1}(feat)$. Once computed, the attention features guide the outputs of various dilated convolutions in the following manner - 
\begin{equation*}
    f_{conv} =  \{f_{i} + (f_{i} \times \sigma(B_{1}(\uparrow f_{attention})))\} \quad
     \text{where} \quad i \in \{6, 10, 14, 18, 22\}
\end{equation*}

where $f_{i} = G_{i}(feat)$ indicate the dilated convolution features, $\uparrow$ indicates bilinear upsampling, and $\sigma$ indicates a sigmoid operation. To influence the convolutional features, the attention features are upsampled to the same spatial resolution, then reducing their channels to $1$ and, finally, passing through a sigmoid layer. These logits are multiplied with the convolutional features, thus filtering crucial per-channel information and inducing an input-dependent nature. Finally, all the refined features are concatenated $\{\}$ in the channel dimension. Thus, the output of an AGLRFE is obtained by concatenating the newly generated convolution features and $feat$ and passing these features through a max pooling \textit{MaxPool} layer followed by a series of convolution layers.

\begin{equation*}
    f_{lr} = B_{1}(MaxPool(\{f_{conv}, feat\}))
\end{equation*}

\subsection{Attention-enhanced Local Processing Module (ALPM)}
The Local Processing Module (LPM) in \cite{dulam2023sodawidenet} employs $3 \times 3$ convolutions to extract local features crucial for precise saliency predictions. This module uses a series of max-pooling layers to identify discriminative features from small neighborhoods, enhancing the model's ability to focus on relevant details. To augment this structure with an added layer of input-specific adaptability, we have refined the LPM by incorporating a Self-Attention mechanism. This Self-Attention is strategically applied to the feature map with the smallest spatial resolution within the LPM, enabling the module to obtain input characteristics from the most informative feature map. 

\begin{gather*}
    fx = MAXPOOL(X) \\
    feat = B_{1}(MAXPOOL(fx)) \\
    fy = B_{1}(\{\uparrow (feat + \textbf{Attn}(feat)), fx\}) \\
    f_{sr} = B_{1}(fx) + fy
\end{gather*}
Above, we illustrate the series of operations in LPM along with the modification highlighted in \textbf{bold}.

\subsection{Loss Function}

\begin{figure*}[ht!]
    \centering
    \begin{tikzpicture}[
 image/.style = {text width=0.165\linewidth,
                 inner sep=0pt, outer sep=0pt},
node distance = 0mm and 0mm
                        ]

\node [image] (frame11)
    {\includegraphics[width=\linewidth]{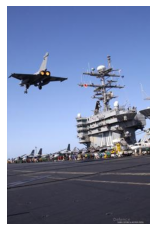}};
\node [image, right=of frame11] (frame21) 
    {\includegraphics[width=\linewidth]{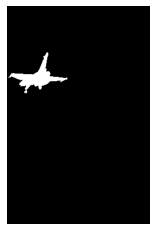}};
\node [image, right=of frame21] (frame31) 
    {\includegraphics[width=\linewidth]{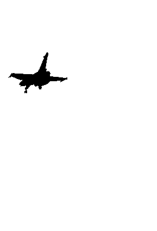}};
\node [image, right=of frame31] (frame41) 
    {\includegraphics[width=\linewidth]{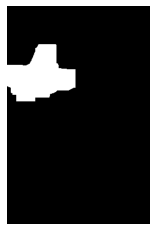}};
\node [image, right=of frame41] (frame51) 
    {\includegraphics[width=\linewidth]{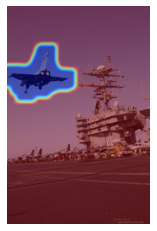}};
\node [image, right=of frame51] (frame61) 
    {\includegraphics[width=\linewidth]{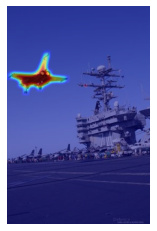}};

\node[below=of frame11]{\tiny I};
\node[below=of frame21]{\tiny FG};
\node[below=of frame31]{\tiny BG($\alpha_{ij}^{bg}$)};
\node[below=of frame41]{\tiny $\alpha_{ij}^{fg}$};
\node[below=of frame51]{\tiny $\alpha_{ij}^{fg}$ SI};
\node[below=of frame61]{\tiny $\alpha_{ij}^{bg}$ SI};

\end{tikzpicture}
    \caption{In the above figure, we visually illustrate the pixels that receive a higher weight for loss computation as shown in equation \ref{fgbgloss}. The third image illustrates in white the pixels that receive a higher weight for the background loss, and the fourth image illustrates the pixels receiving a higher weight for the foreground loss. The last two images depict the important pixels (in blue) superimposed (SI) on the input image.}
    \label{fig:AlphaBetaImage}
\end{figure*}
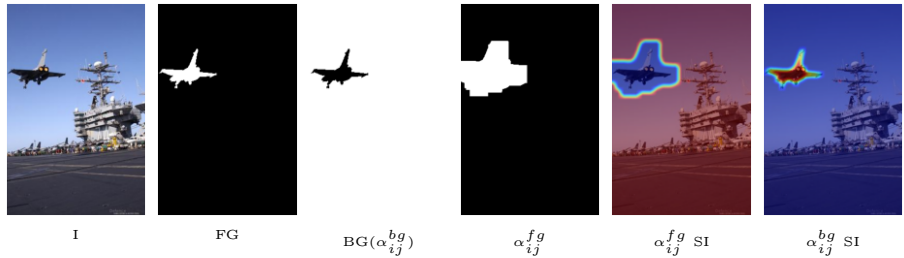

To train the proposed model, we create a custom loss that directs the model to produce accurate saliency predictions. Following \cite{dulam2023sodawidenet}, our model has saliency and contour outputs, supervised with the ground-truth saliency and contour maps, as shown in Equations \ref{TotalSalLoss} and \ref{TotalConLoss}, respectively.

\begin{equation}\label{TotalSalLoss}
\begin{split}
    L_{salient} = \sum_{i=1}^{2} (L_{AGLRFE_{(i)}}^{sal} + L_{ALPM_{(i)}}^{sal} + L_{CFM_{(i)}}^{sal} \\ + L_{MRFFAM_{(i)}}^{sal} + L_{CFMD_{(i)}}^{sal})
\end{split}
\end{equation}


\begin{equation}\label{TotalConLoss}
\begin{split}
    L_{contour} = L_{MRFFAM_{(1)}}^{con} + L_{MRFFAM_{(2)}}^{con} \\ + L_{CFMD_{(1)}}^{con} + L_{CFMD_{(2)}}^{con} \\
    L^{con} = 0.001 \cdot L_{BCE} + L_{dice} \quad \quad \quad
\end{split}
\end{equation}

where $L_{BCE}$ and $L_{dice}$ are the Binary Cross Entropy and Dice loss, respectively. Unlike other methods, we supervise foreground and background saliency maps. Hence, $L^{sal}$ is divided into two parts. The foreground loss denoted by $L_{fg}$ is the same as the loss function in SODAWideNet, whereas $L_{bg}$ is the newly added background supervision. For COCO pre-training, we set $\beta$ to one, and for DUTS, we set it to 0.5.

\begin{equation}\label{fgbgloss}
    \begin{split}
    L^{sal} = L_{fg} + \beta \cdot L_{bg} \quad \quad \quad \quad \\
        L_{fg} = \alpha^{fg} \cdot L_{bce} + \alpha^{fg} \cdot L_{iou} + \alpha^{fg} \cdot L_{1} \\
        L_{bg} = \alpha^{bg} \cdot L_{bce} + \alpha^{bg} \cdot L_{iou} + \alpha^{bg} \cdot L_{1}
    \end{split}
\end{equation}

where $L_{bce}$, $L_{iou}$, and $L_{1}$ are per-pixel Binary Cross Entropy, Intersection-over-Union, and L1 losses, respectively. $\alpha_{ij}^{fg}$ and $\alpha_{ij}^{bg}$ are per-pixel weights determined for each pixel during foreground and background supervision. Per-pixel weights ensure that wrong predictions for specific pixels are penalized heavily. Below, we illustrate the procedure to compute these values. Both $\alpha^{fg}$ and $\alpha^{bg}$ have the exact spatial resolution as the output. Hence, we compute pixel-wise loss in Equation \ref{fgbgloss} and multiply them with $\alpha$. 
\begin{gather*}
    \alpha_{ij}^{fg} = max(GT_{ij})_{31 \times 31} \\
    \alpha_{ij}^{bg} = BackgroundGT_{ij} \\
    BackgroundGT = \mathds{1}\ - GT
\end{gather*}
$\alpha_{ij}^{fg}$ for a pixel at spatial location $(i, j)$ is calculated by finding the largest value in a $31 \times 31$ window centered on that pixel location. Whereas $\alpha_{ij}^{bg}$ equals the pixel's intensity value in the background map. Figure \ref{fig:AlphaBetaImage} indicates the values of $\alpha_{ij}^{fg}$ and $\alpha_{ij}^{bg}$ superimposed on the input. Finally, from Equations \ref{TotalSalLoss}, and \ref{TotalConLoss}, the total loss to train our model is given as

\begin{equation} \label{TotalLoss}
    L_{total} = L_{salient} + L_{contour} 
\end{equation}

\section{Experiments and Results}

\subsection{Datasets}
We pre-train our model on the modified COCO dataset of 118K annotated images and ground truth pairs. We further augment it using horizontal and vertical flipping, taking the pre-training data size to 354K images. Then, we fine-tune our model on the DUTS~\cite{wang2017DUTS-TE} dataset, containing 10,553 images for training. We augment the data using horizontal and vertical flipping to obtain a training dataset of 31,659 images. We use five datasets to evaluate the proposed model. They are DUTS-Test\cite{wang2017DUTS-TE} consisting of 5019 images, DUT-OMRON\cite{yang2013saliencyOMRON} which consists of 5168 images, HKU-IS\cite{LiYu15HKU} which consists of 4447 images, ECSSD\cite{shi2015hierarchicalECSSD} which consists of 1000 images and PASCAL-S\cite{li2014secretsPASCAL-S} dataset consisting of 850 images.

\begin{table*}[ht!]
  \begin{adjustbox}{width=\linewidth}
  \begin{tabular}{|cc|ccccc|ccccc|ccccc|}
    \hline
      Method& Params. (M)&  \multicolumn{5}{c|}{DUTS-TE~\cite{wang2017DUTS-TE}} & \multicolumn{5}{c|}{DUT-OMRON~\cite{yang2013saliencyOMRON}} & \multicolumn{5}{c|}{HKU-IS~\cite{LiYu15HKU}} \\
    \hline
     & & $F_{max}$ & MAE  & $S_{m}$& $E_{m}$& $F_{w}$ &
     $F_{max}$ & MAE & $S_{m}$& $E_{m}$& $F_{w}$ & 
    $F_{max}$ & MAE &  $S_{m}$& $E_{m}$& $F_{w}$  \\
    \hline
    PiCAN$_{\text{CVPR'18}}$~\cite{liu2018picanet}&47.22&
    0.860&0.051&0.869&0.862&0.755&
    0.803&0.065&0.832&0.841&0.695&
    0.918&0.043&0.904&0.936&0.840\\
    
    BASN$_{\text{CVPR'19}}$~\cite{qin2019basnet}&87.06&
    0.860&0.048&0.866&0.884&0.803&
    0.805&0.056&0.836&0.861&0.751&
    0.928&0.032&0.909&0.946&0.889\\
    
    F3-N$_{\text{AAAI'20}}$~\cite{wei2020f3net} & 26.5&
    0.891&0.035&0.888&0.902&0.835&
    0.813&0.053&0.838&0.870&0.747&
     0.937&0.028&0.917&0.953&0.900\\

    LDF$_{\text{CVPR'20}}$~\cite{ldf}& 25.15&
    0.898& 0.034&0.892 & 0.910&0.845&
    0.820&0.051&0.838 & 0.873&0.752&
    0.939&0.027&0.919&0.954&0.904\\

    PA-KRN$_{\text{AAAI'21}}$~\cite{pakrn} & 68.68&
    0.907& 0.033&0.900 & 0.916&0.861&
    0.834&0.050&0.853 & 0.885&0.779&
    0.943&0.027&0.923 &0.955&0.909 \\
    
    VST$_{\text{ICCV'21}}$~\cite{Liu_2021_ICCV_VST} & 44.48 &
    0.890&  0.037&0.896 & 0.892&0.828&
    0.825&0.058&0.850 & 0.861&0.755&
    0.942&0.029&0.928&0.953&0.897\\

    PSG$_{\text{TIP'21}}$~\cite{yang2021progressivepsg}&25.55&
    0.886&0.036&0.883&0.908&0.835&
    0.811&0.052&0.831&0.870&0.747&
    0.938&0.027&0.919&0.906&0.958\\
    
    ET$_{\text{NeurIPS'21}}$~\cite{jing_ebm_sod21generative}&118.96&
    0.910&0.029&0.909&0.918&0.871&
    0.839&0.050&0.858&0.886&0.788&
    0.947&0.023&0.930&0.961&0.920\\

    RCSB$_{\text{WACV'22}}$~\cite{Ke_2022_WACV_RCSB} & 27.90& 
    0.889&0.035&0.878&0.903&0.840&
    0.810&0.045&0.820&0.856&0.723&
    0.938&0.027&0.918&0.954&0.909\\
    
    CSF-R2$_{\text{TPAMI'22}}$~\cite{gao2020sod100kcsf}& 36.53&
    0.890&0.037&0.890&0.897&0.823&
     0.815&0.055&0.838&0.861&0.734&
    0.935&0.030&0.921&0.952&0.891\\
    
    EDN$_{\text{TIP'22}}$~\cite{wu2022edn} & 42.85 & 
    0.895&0.035&0.892&0.908&0.845&
    0.828&0.048&0.846&0.876&0.770&
    0.941&0.026&0.924&0.956&0.908\\
    
    PGN$_{\text{CVPR'22}}$~\cite{xie2022pyramid} & 72.70 &
    0.917&0.027&0.911&0.922&0.874&
    0.835&0.045&0.855&0.887&0.775&
    0.948&0.024&0.929&0.961&0.916\\
    
    

    ICON-R$_{\text{TPAMI'22}}$~\cite{zhuge2021salienticon} & 33.09&
    0.892&0.037&0.889 & 0.902&0.837&
    0.825&0.057&0.844& 0.870&0.761&
    0.939&0.029&0.920&0.952&0.902 \\

    TR5$_{\text{AAAI'22}}$~\cite{lee2022tracer} & 31.30&
    0.916& 0.026&0.909 & 0.927&0.883&
    0.834&0.042&0.847 &0.880&0.787&
    0.947&0.022&0.930&0.961&0.922 \\

    SR$_{\text{TMM'23}}$~\cite{SelfReformer} & 220&
    0.925& 0.024&0.921 &0.924&0.886&
    0.838&0.043&0.859 & 0.884&0.782&
    0.951&0.023&0.934&0.962&0.918 \\

    RMF$_{\text{ACMMM'23}}$~\cite{rmformer} & 87.52&
    0.931& 0.023&0.925 &0.933&0.900&
    0.861&0.040&0.877 & 0.904&0.819&
    0.957&0.019&0.940&0.968&0.934 \\
    VSC$_{\text{CVPR'24}}$~\cite{luo2024vscode} & 74.72&
    0.931& 0.024&0.926 &0.931&0.897&
    0.861&0.042&0.876 & 0.899&0.813&
    0.957&0.021&0.940 &0.965&0.930\\

    \hline

    \textbf{Ours} & \textbf{26.58}&
    \textbf{0.917}& \textbf{0.029}&\textbf{0.910} & \textbf{0.916}&\textbf{0.870}&
    \textbf{0.848}&\textbf{0.045}&\textbf{0.868} & \textbf{0.896}&\textbf{0.796}&
    \textbf{0.950}&\textbf{0.024}&\textbf{0.932} &\textbf{0.961}&\textbf{0.917}\\

    \textbf{Ours-M} & \textbf{6.66}&
    \textbf{0.901}& \textbf{0.035}&\textbf{0.898} & \textbf{0.907}&\textbf{0.846}&
    \textbf{0.844}&\textbf{0.048}&\textbf{0.861} & \textbf{0.888}&\textbf{0.784}&
    \textbf{0.949}&\textbf{0.025}&\textbf{0.932}&\textbf{0.960}&\textbf{0.915} \\

    \textbf{Ours-S} & \textbf{1.67}&
    \textbf{0.887}& \textbf{0.039}&\textbf{0.886} & \textbf{0.898}&\textbf{0.824}&
    \textbf{0.834}&\textbf{0.051}&\textbf{0.854} & \textbf{0.882}&\textbf{0.772}&
    \textbf{0.941}&\textbf{0.028}&\textbf{0.925}&\textbf{0.955}&\textbf{0.904}\\
    
    \hline
  \end{tabular}
  \end{adjustbox}
  \caption{Comparison of our method with 17 other methods in terms of max F-measure $F_{max}$, MAE, $S_{m}$, $E_{m}$, and $F_{w}$ on DUTS-TE, DUT-OMRON, and HKU-IS datasets.}
  \label{tab:ResultsTable1}
\end{table*}

\begin{table*}[ht!]
  \centering
  \begin{adjustbox}{width=0.8\linewidth}
  \begin{tabular}{|cc|ccccc|ccccc|}
    \hline
      Method& Params. (M) & \multicolumn{5}{c|}{ECSSD~\cite{shi2015hierarchicalECSSD}} & \multicolumn{5}{c|}{PASCAL-S~\cite{li2014secretsPASCAL-S}}\\
    \hline
     & &
    $F_{max}$ & MAE &  $S_{m}$  & $E_{m}$  &$F_{w}$  &
    $F_{max}$ & MAE &  $S_{m}$ &$E_{m}$  &$F_{w}$  \\
    \hline
    PiCAN$_{\text{CVPR'18}}$~\cite{liu2018picanet}&47.22&
    0.935&0.046&0.917&0.913&0.867&
    0.868&0.078&0.852&0.837&0.779\\
    
    BASN$_{\text{CVPR'19}}$~\cite{qin2019basnet}&87.06&
    0.942&0.037&0.916&0.921&0.904&
    0.860&0.079&0.834&0.850&0.797\\
    
    F3-N$_{\text{AAAI'20}}$~\cite{wei2020f3net} & 26.5&
     0.945&0.033&0.924&0.927&0.912&
    0.882&0.064&0.857&0.863&0.823\\

    LDF$_{\text{CVPR'20}}$~\cite{ldf} & 25.15&
    0.950&0.034&0.924 & 0.925&0.915&
    0.887&0.062&0.859&0.869&0.829\\

    PA-KRN$_{\text{AAAI'21}}$~\cite{pakrn} & 68.68&
    0.953&0.032&0.928 & 0.924&0.918&
    -&-&-&-&-\\
    
    VST$_{\text{ICCV'21}}$~\cite{Liu_2021_ICCV_VST} & 44.48 &
    0.951&0.033&0.932&0.918&0.910&
    0.890&0.062&0.871&0.846&0.827\\

    PSG$_{\text{TIP'21}}$~\cite{yang2021progressivepsg}&25.55&
    0.949&0.031&0.925&0.928&0.917&
    0.886&0.063&0.858&0.863&0.830\\
    
    ET$_{\text{NeurIPS'21}}$~\cite{jing_ebm_sod21generative}&118.96&
    0.959&0.023&0.942&0.933&0.937&
    0.900&0.055&0.876&0.869&0.863\\

    RCSB$_{\text{WACV'22}}$~\cite{Ke_2022_WACV_RCSB} & 27.90&
    0.944&0.033&0.921&0.923&0.916&
    0.886&0.061&0.857&0.858&0.834\\
    
    CSF-R2$_{\text{TPAMI'22}}$~\cite{gao2020sod100kcsf}& 36.53&
    0.950&0.033&0.930&0.928&0.910&
    0.886&0.069&0.862&0.855&0.818\\
    
    EDN$_{\text{TIP'22}}$~\cite{wu2022edn} & 42.85 &
    0.951&0.032&0.927&0.929&0.918&
    0.891&0.065&0.860&0.867&0.832\\
    
    PGN$_{\text{CVPR'22}}$~\cite{xie2022pyramid} & 72.70 &
    0.960&0.027&0.918&0.932&0.929&
    0.904&0.056&0.874&0.878&0.849\\
    
    ICON-R$_{\text{TPAMI'22}}$~\cite{zhuge2021salienticon} & 33.09&
    0.950&0.032&0.929 & 0.929& 0.918&
    0.888&0.066&0.860&0.861&0.828\\
    
    TR5$_{\text{AAAI'22}}$~\cite{lee2022tracer} & 31.30&
    0.956&0.027&0.933 & 0.926 & 0.931&
    0.907&0.051&0.878&0.875&0.859\\

    SR$_{\text{TMM'23}}$~\cite{SelfReformer} & 220&
    0.962&0.025&0.941 & 0.935 & 0.932 &
    -&-&-&-&-\\

    RMF$_{\text{ACMMM'23}}$~\cite{rmformer} & 87.52&
    0.964&0.020&0.947 &0.938&0.946&
    -&-&-&-&-\\

    VSC$_{\text{CVPR'24}}$~\cite{luo2024vscode} & 74.72&
    0.965&0.021&0.949 & 0.934 & 0.942 & 
    0.912&0.051&0.885&0.870&0.863\\

    \hline

    \textbf{Ours} & \textbf{26.58}&
   \textbf{ 0.957}&\textbf{0.029}&\textbf{0.935} & \textbf{0.927} & \textbf{0.922} &
    \textbf{0.901}&\textbf{0.062}&\textbf{0.875} & \textbf{0.86}8 & \textbf{0.844}\\

    \textbf{Ours-M} & \textbf{6.66}&
    \textbf{0.952}&\textbf{0.033}&\textbf{0.930} & \textbf{0.26} & \textbf{0.915} &
    \textbf{0.895}&\textbf{0.065}&\textbf{0.866} &\textbf{ 0.865 }& \textbf{0.834}\\

    \textbf{Ours-S} & \textbf{1.67}&
    \textbf{0.946}&\textbf{0.037}&\textbf{0.923} & \textbf{0.924} & \textbf{0.906} &
    \textbf{0.884}&\textbf{0.070}&\textbf{0.857} & \textbf{0.860} & \textbf{0.818}\\
    
    \hline
  \end{tabular}
  \end{adjustbox}
  \caption{Comparison of our method with 17 other methods in terms of max F-measure $F_{max}$, MAE, $S_{m}$, $E_{m}$, and $F_{w}$ measures on ECSSD and PASCAL-S datasets.}
  \label{tab:ResultsTable2}
\end{table*}

\begin{figure*}[ht!]
    \centering
    \begin{tikzpicture}[
 image/.style = {text width=0.098\linewidth,
                 inner sep=0pt, outer sep=0pt},
node distance = 0mm and 0mm
                        ] 
\node [image] (frame11)
   {\includegraphics[width=\linewidth]{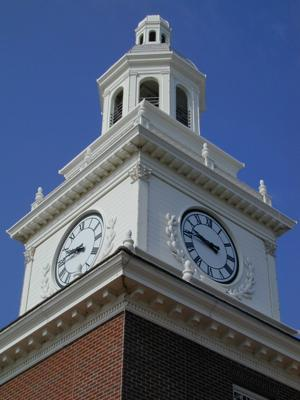}};
\node [image,right=of frame11] (frame21) 
    {\includegraphics[width=\linewidth]{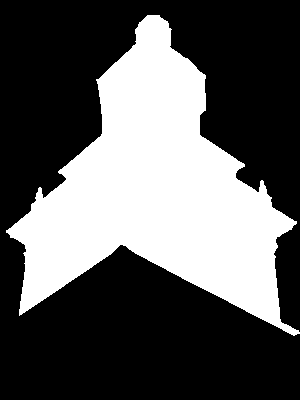}};
\node[image,right=of frame21] (frame31)
    {\includegraphics[width=\linewidth]{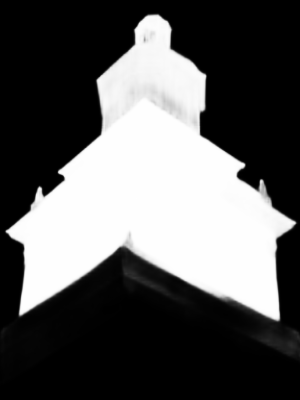}};
\node[image,right=of frame31] (frame41)
    {\includegraphics[width=\linewidth]{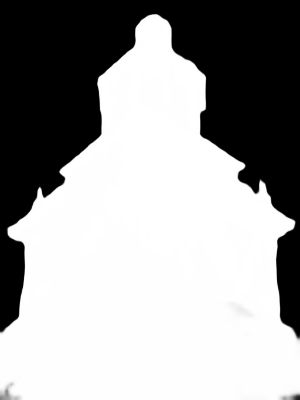}};
\node[image,right=of frame41] (frame51)
    {\includegraphics[width=\linewidth]{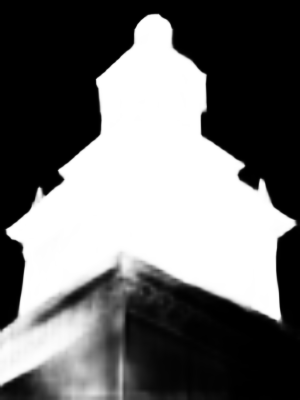}};
\node[image,right=of frame51] (frame61)
    {\includegraphics[width=\linewidth]{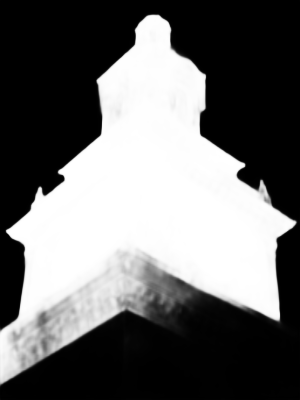}};
\node[image,right=of frame61] (frame71)
    {\includegraphics[width=\linewidth]{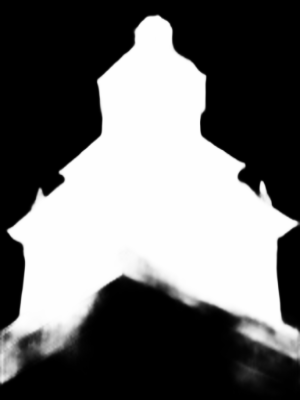}};
\node[image,right=of frame71] (frame81)
    {\includegraphics[width=\linewidth]{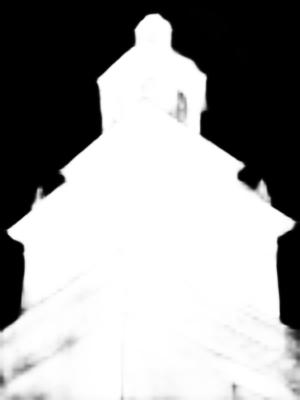}};
\node[image,right=of frame81] (frame91)
    {\includegraphics[width=\linewidth]{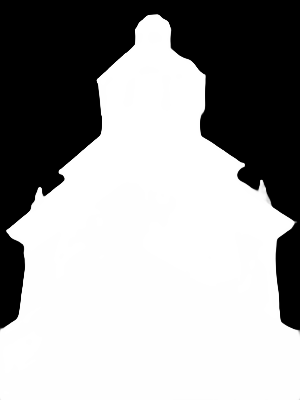}};
\node[image,right=of frame91] (frame101)
    {\includegraphics[width=\linewidth]{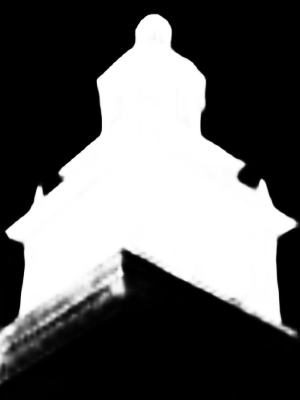}};

\node [image, below=of frame11] (frame12)
   {\includegraphics[width=\linewidth]{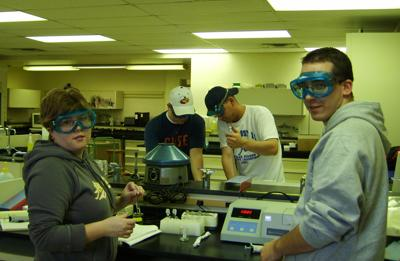}};
\node [image,right=of frame12] (frame22) 
    {\includegraphics[width=\linewidth]{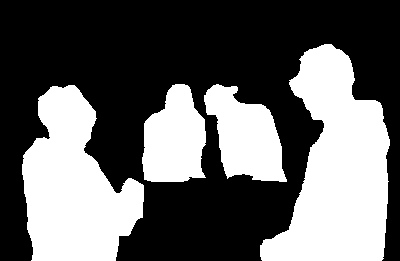}};
\node[image,right=of frame22] (frame32)
    {\includegraphics[width=\linewidth]{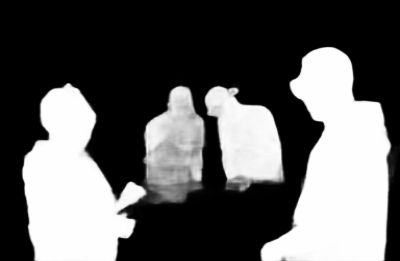}};
\node[image,right=of frame32] (frame42)
    {\includegraphics[width=\linewidth]{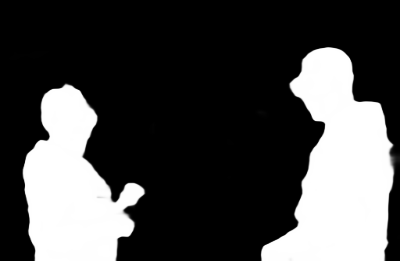}};
\node[image,right=of frame42] (frame52)
    {\includegraphics[width=\linewidth]{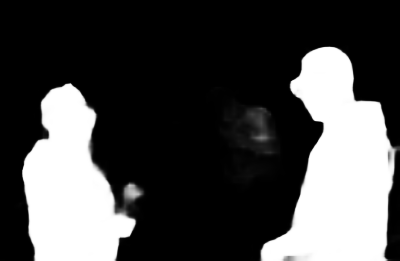}};
\node[image,right=of frame52] (frame62)
    {\includegraphics[width=\linewidth]{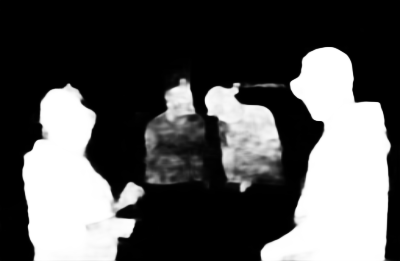}};
\node[image,right=of frame62] (frame72)
    {\includegraphics[width=\linewidth]{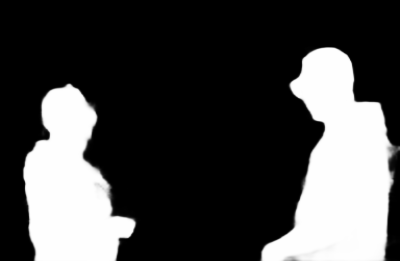}};
\node[image,right=of frame72] (frame82)
    {\includegraphics[width=\linewidth]{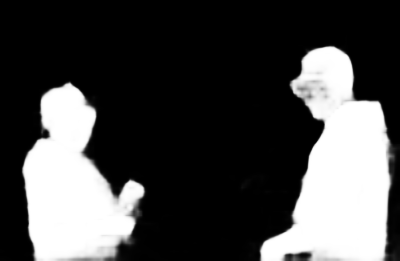}};
\node[image,right=of frame82] (frame92)
    {\includegraphics[width=\linewidth]{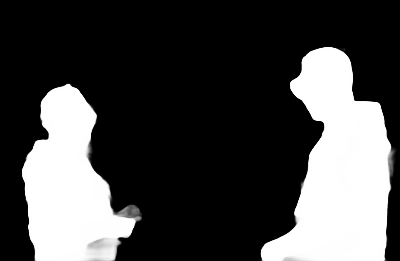}};
\node[image,right=of frame92] (frame102)
    {\includegraphics[width=\linewidth]{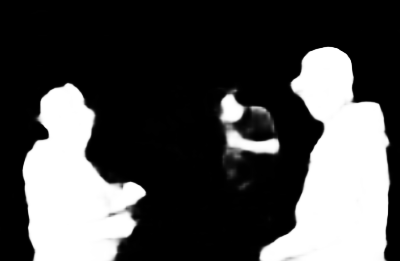}};

\node [image, below=of frame12] (frame13)
   {\includegraphics[width=\linewidth]{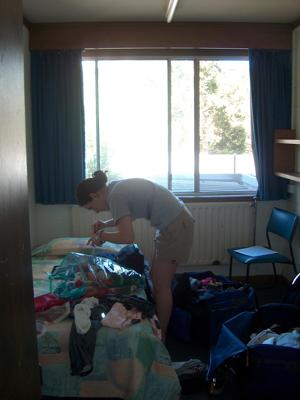}};
\node [image,right=of frame13] (frame23) 
    {\includegraphics[width=\linewidth]{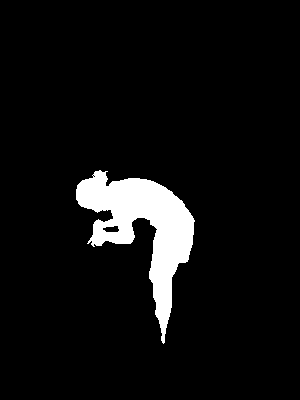}};
\node[image,right=of frame23] (frame33)
    {\includegraphics[width=\linewidth]{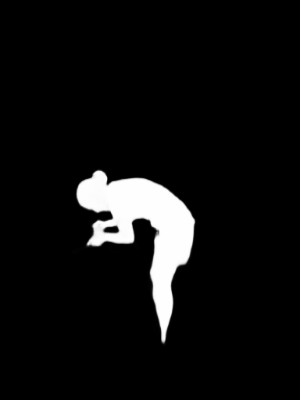}};
\node[image,right=of frame33] (frame43)
    {\includegraphics[width=\linewidth]{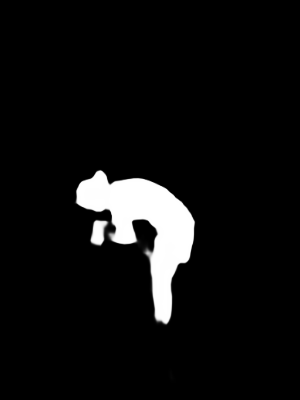}};
\node[image,right=of frame43] (frame53)
    {\includegraphics[width=\linewidth]{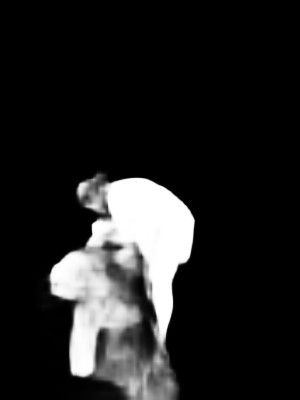}};
\node[image,right=of frame53] (frame63)
    {\includegraphics[width=\linewidth]{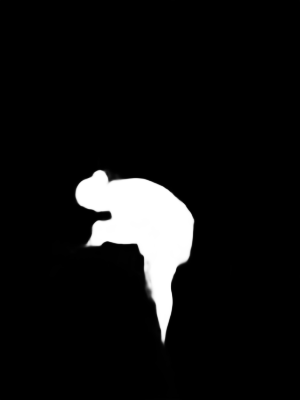}};
\node[image,right=of frame63] (frame73)
    {\includegraphics[width=\linewidth]{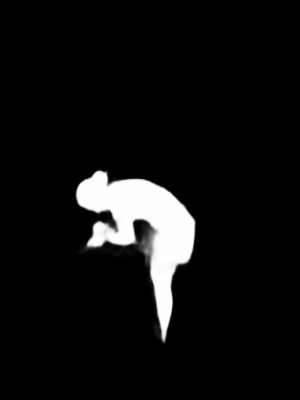}};
\node[image,right=of frame73] (frame83)
    {\includegraphics[width=\linewidth]{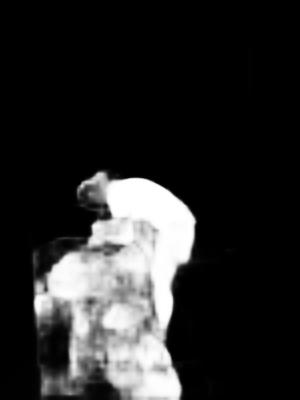}};
\node[image,right=of frame83] (frame93)
    {\includegraphics[width=\linewidth]{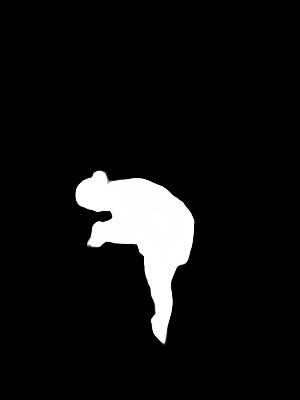}};
\node[image,right=of frame93] (frame103)
    {\includegraphics[width=\linewidth]{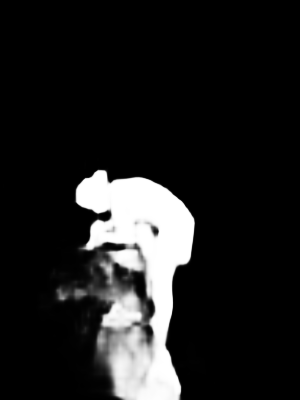}};

\node [image, below=of frame13] (frame14)
   {\includegraphics[width=\linewidth]{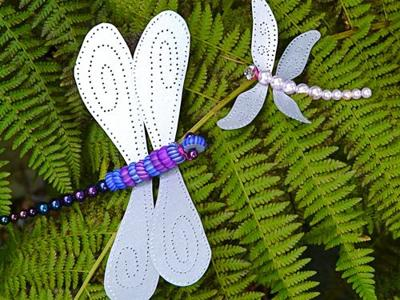}};
\node [image,right=of frame14] (frame24) 
    {\includegraphics[width=\linewidth]{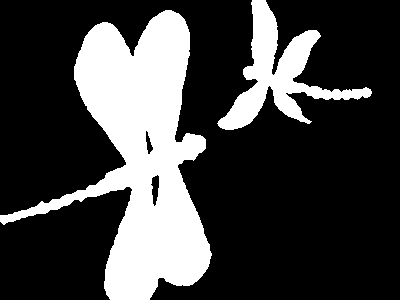}};
\node[image,right=of frame24] (frame34)
    {\includegraphics[width=\linewidth]{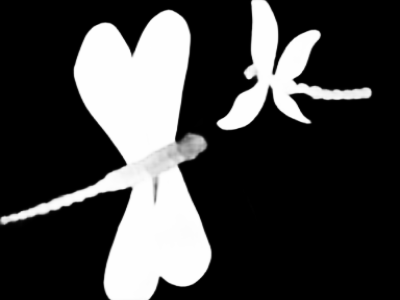}};
\node[image,right=of frame34] (frame44)
    {\includegraphics[width=\linewidth]{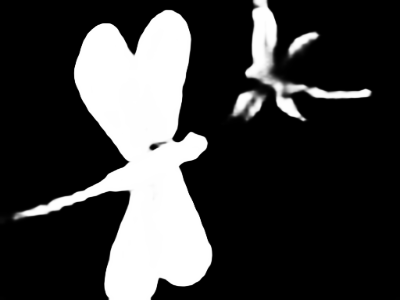}};
\node[image,right=of frame44] (frame54)
    {\includegraphics[width=\linewidth]{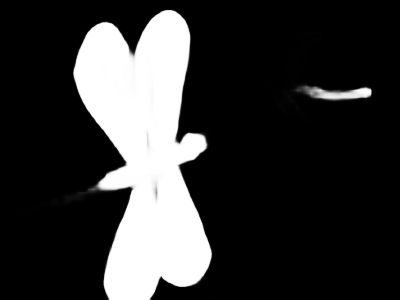}};
\node[image,right=of frame54] (frame64)
    {\includegraphics[width=\linewidth]{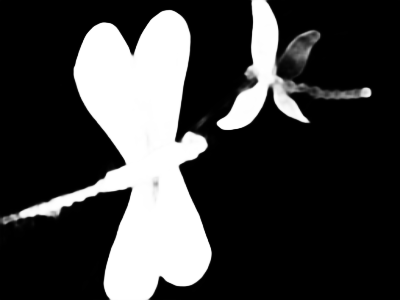}};
\node[image,right=of frame64] (frame74)
    {\includegraphics[width=\linewidth]{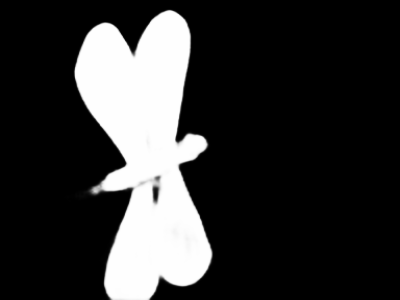}};
\node[image,right=of frame74] (frame84)
    {\includegraphics[width=\linewidth]{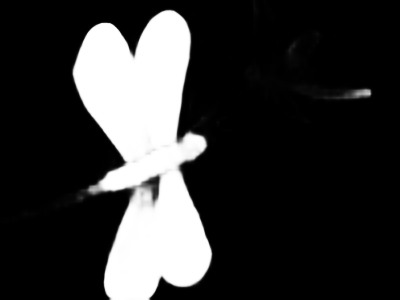}};
\node[image,right=of frame84] (frame94)
    {\includegraphics[width=\linewidth]{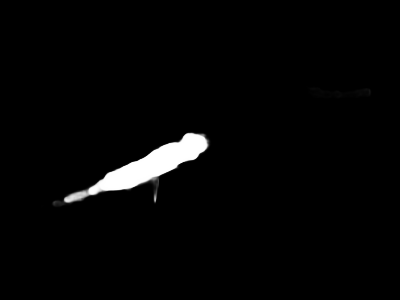}};
\node[image,right=of frame94] (frame104)
    {\includegraphics[width=\linewidth]{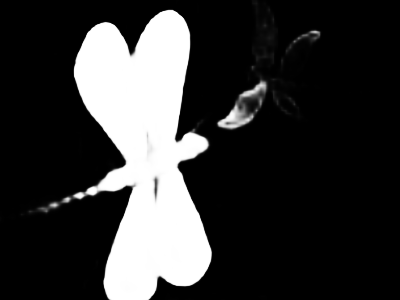}};


\node[below=of frame14]{\tiny I};
\node[below=of frame24]{\tiny GT};
\node[below=of frame34]{\tiny OURS};
\node[below=of frame44]{\tiny \cite{lee2022tracer}};
\node[below=of frame54]{\tiny \cite{zhuge2021salienticon}};
\node[below=of frame64]{\tiny \cite{xie2022pyramid}};
\node[below=of frame74]{\tiny \cite{wu2022edn}};
\node[below=of frame84]{\tiny \cite{gao2020sod100kcsf}};
\node[below=of frame94]{\tiny \cite{Ke_2022_WACV_RCSB}};
\node[below=of frame104]{\tiny \cite{jing_ebm_sod21generative}};

\end{tikzpicture}
    \caption{In the above figure, we visually compare our results against other models.}
    \label{fig:VisualResults}
\end{figure*}

\subsection{Implementation Details}
For COCO pre-training, we train our model for 21 epochs. The LR is set at 0.001 and multiplied by 0.5 after 15 epochs. For SOD fine-tuning, we train our model for a further 11 epochs with the same LR as the COCO stage. LR is multiplied by 0.1 after five epochs. We follow the procedure used in U-Net\cite{ronneberger2015UNET} to initialize the weights of our model. Images are resized to $384 \times 384$ for training and testing. The predictions for the background saliency supervision are generated by multiplying the pre-sigmoid predictions with $-1$, thus turning the negative values into positive and positive values into negative. The evaluation metrics for comparing our works with prior works are the Mean Absolute Error(MAE), maximum F-measure, the S-measure\cite{smeasure}, the E-measure \mbox{\cite{fan2018enhancedEmeasure}}, and the weighted F-measure. The lower the MAE and the higher the $F_{max}$, $S_{m}$, $E_{m}$, and $F_{w}$ scores, the better the model.


 


\subsection{Quantitative and Qualitative Results}

Tables \ref{tab:ResultsTable1} and \ref{tab:ResultsTable2} detail the quantitative performance of SODAWideNet++ compared to 17 other state-of-the-art models. Notably, on the DUTS-TE, DUT-OMRON, and HKU-IS datasets, our model achieves competitive scores in  $F_{max}$, $S_{m}$, and $E_{m}$ measures which signify highly confident and accurate predictions. Especially while SODAWideNet++ (Ours) uses significantly fewer parameters (\textbf{33\%}, \textbf{35\%}, and \textbf{13\%} of trainable parameters compared to the state-of-the-art RMF \cite{rmformer}, VSC \cite{luo2024vscode}, and SR \cite{SelfReformer}, respectively). Additionally, the smaller models SODAWideNet++-M (Ours-M) perform considerably well and surpass the older state-of-the-art models such as VST and RCSB. Similarly, the smallest model, SODAWideNet++-S (Ours-S), also illustrates great performance and can be useful in parameter-constrained situations.

The visual results, as depicted in Figure \ref{fig:VisualResults}, further substantiate the robustness of SODAWideNet++ across diverse scenarios, including images with large foreground objects, scenes containing multiple objects (notably in the second and fourth rows), and environments characterized by complex backgrounds (third and fourth rows). These results highlight the model's ability to detect salient objects in challenging visual conditions.

\section{Ablation Experiments}

Through ablation experiments, we delve into the effects of the proposed design choices on the proposed model. All reported numbers are on the DUTS test split.

\subsection{SODAWideNet vs. SODAWideNet++}

We performed a comparative analysis of SODAWideNet and SODAWideNet++ performance, evaluating both models from scratch and after pre-training on the modified COCO dataset, ensuring comparisons were made against models of similar size for consistency. SODAWideNet++ integrates the Multi-Receptive Field Feature Aggregation Module (MRFFAM) and Multi-scale Attention (MSA) into a single module, the Attention Guided Long Range Feature Extraction (AGLRFE). This integration reduced redundancy when extracting long-range features. Additionally, incorporating background supervision in SODAWideNet++ enhances its ability to distinguish between foreground and background areas, leading to more precise saliency results. Table \ref{tab:sodawidenetvssodawidenet++} illustrates the quantitative performance.

\begin{table}[h!]
    \centering
    \begin{tabular}{|c|c|c|c|c|}
        \hline
         \textbf{Pre-training mechanism} & \textbf{Model} & Size (in M) & \textbf{$F_{max}$} & \textbf{MAE}  \\
         \hline
         Scratch & SODAWideNet & 9.03 & 0.883 & 0.043 \\
         \hline
         Mod. COCO & SODAWideNet & 9.03 & 0.899& 0.035\\
         \hline
         Scratch & SODAWideNet++ & 6.66& 0.881 & 0.043 \\
         \hline
         Mod. COCO & SODAWideNet++ & 6.66 & 0.901 & 0.035\\
         \hline
    \end{tabular}
    \caption{SODAWideNet vs. SODAWideNet++.}
    \label{tab:sodawidenetvssodawidenet++}
\end{table}




\subsection{ImageNet vs. COCO Pre-training}

Table~\ref{tab:cocovsimagenet} compares our proposed models' performance when pre-trained on either the ImageNet or the modified COCO dataset. The results show an improvement with COCO pre-training, where the model experiences a 1.2\% increase in performance compared to ImageNet. This significant enhancement can be attributed to the fact that ImageNet pre-training tends to separate the development of the backbone and the feature refinement modules during the pre-training phase, leading to a disconnect in feature interpretation between the encoding and decoding stages when fine-tuning for SOD. Also, training the model from scratch achieves decent performance, highlighting the effectiveness of incorporating self-attention into a convolutional neural network.


\begin{table}[h!]
    \centering
    \begin{tabular}{|c|c|c|c|}
        \hline
         \textbf{Pre-training mechanism} & \textbf{Model} & \textbf{$F_{max}$} & \textbf{MAE}  \\
         \hline
         Scratch & SODAWideNet++ & 0.890 & 0.039 \\
         \hline
         ImageNet & SODAWideNet++ & 0.905& 0.036\\
         \hline
         Mod. COCO & SODAWideNet++ & 0.917& 0.029\\
         \hline
    \end{tabular}
    \caption{ImageNet vs. COCO vs. Training from scratch.}
    \label{tab:cocovsimagenet}
\end{table}

\subsection{Pre-training another model using the modified COCO dataset}

We pre-train the PGN \mbox{\cite{xie2022pyramid}} model using the modified COCO dataset without their ImageNet pre-trained weights. Tables \mbox{\ref{tab:pgnet1}} and \mbox{}{\ref{tab:pgnet2}} contain a quantitative evaluation of the COCO pre-trained PGN, ImageNet pre-trained PGN, and PGN trained from scratch on the DUTS dataset. We have included our model for comparison. Pre-training using the COCO dataset significantly outperforms training from scratch and delivers competitive results against the ImageNet pre-trained PGN model without optimal hyperparameters. 

\begin{table*}[ht!]
  \begin{adjustbox}{width=\linewidth}
  \begin{tabular}{|c|c|ccccc|ccccc|ccccc|}
    \hline
      Method& Params. (M)&  \multicolumn{5}{c|}{DUTS-TE} & \multicolumn{5}{c|}{DUT-OMRON} & \multicolumn{5}{c|}{HKU-IS} \\
    \hline
     & & $F_{max}$ & MAE  & $S_{m}$ & $E_{m}$ & $E_{m}$ & $F_{max}$ & MAE & $S_{m}$ & $E_{m}$ & $E_{m}$ &
    $F_{max}$ & MAE &  $S_{m}$ & $E_{m}$ &$E_{m}$  \\
    \hline

    PGN-S & 72.70&
    0.823& 0.060&0.833 & 0.851& 0.731&
    0.779&0.068&0.809 & 0.837&0.690&
    0.909&0.042&0.891&0.934&0.852 \\

    PGN$_{\text{CVPR'22}}$~\cite{xie2022pyramid} & 72.70&
    0.917& 0.027&0.911 & 0.922&0.874&
    0.835&0.045&0.855 & 0.887&0.775&
    0.948&0.024&0.929&0.961&0.916 \\

    PGN-COCO$_{\text{CVPR'22}}$~\cite{xie2022pyramid} & 72.70&
    0.882& 0.040&0.879 & 0.896& 0.820&
    0.803&0.057&0.828 & 0.859&0.732&
    0.931&0.031&0.912&0.948&0.732 \\

    \hline

    \textbf{Ours} & \textbf{26.58}&
    \textbf{0.917}& \textbf{0.029}&\textbf{0.910} &\textbf{0.916}&\textbf{0.916} & 
    \textbf{0.848}&\textbf{0.045}&\textbf{0.868} &\textbf{0.896}&\textbf{0.916} & 
    \textbf{0.950}&\textbf{0.024}&\textbf{0.932}
    &\textbf{0.960}&\textbf{0.916} \\
    \hline
  \end{tabular}
  \end{adjustbox}
  \caption{Comparison of pre-training (COCO vs. ImageNet) of PGN method.}
  \label{tab:pgnet1}
\end{table*}

\begin{table*}[ht!]
  \centering
  \begin{adjustbox}{width=0.8\linewidth}
  \begin{tabular}{|c|c|ccccc|ccccc|}
    \hline
      Method& Params. (M) & \multicolumn{5}{c|}{ECSSD} & \multicolumn{5}{c|}{PASCAL-S}\\
    \hline
     & &
    $F_{max}$ & MAE &  $S_{m}$  & $E_{m}$  & $F_{w}$  &
    $F_{max}$ & MAE &  $S_{m}$ &  $E_{m}$ & $F_{w}$  \\
    \hline

    PGN-S & 72.70&
    0.916&0.054&0.891 & 0.907& 0.855 &
    0.839&0.094&0.812&0.824&0.750\\

    PGN$_{\text{CVPR'22}}$~\cite{xie2022pyramid} & 72.70&
    0.960&0.027&0.938 & 0.932& 0.929 &
    0.904&0.056&0.874&0.878 &0.849\\

    PGN-COCO$_{\text{CVPR'22}}$~\cite{xie2022pyramid} & 72.70&
    0.940&0.039&0.913 & 0.915& 0.896 &
    0.878&0.069&0.852&0.864&0.818\\

    \hline

    \textbf{SODAWideNet++} & \textbf{26.58}&
   \textbf{ 0.957}&\textbf{0.029}&\textbf{0.935} & 0.927 & 0.922&
   \textbf{0.901}&\textbf{0.062}&\textbf{0.875}&\textbf{0.870} &0.845\\

    \hline
  \end{tabular}
  \end{adjustbox}
  \caption{Comparison of pre-training (COCO vs. ImageNet) of PGN method.}
  \label{tab:pgnet2}
\end{table*}

\subsection{Ablation experiments corresponding to ALGRFE, ALPM, CFM, \& MRFFAM}

Table \mbox{\ref{tab:components}} provides a quantitative comparison of the impact of each component of the proposed architecture. The absence of AGLRFE (row one) reduces the model's ability to capture long-range dependencies, leading to the lowest performance across all configurations. The removal of ALPM (row two) reduces the model's ability to capture local features. Similarly, removing CFM (row three) causes a similar decline in performance due to the lack of a complex way to integrate local and global features. Also, removing MRFFAM (row four) on the decoder side results in degraded performance, reinforcing the importance of using various receptive fields to decode features. The inclusion of all the four components produces the best model.
\begin{table}[h!]
\centering
  \begin{tabular}{|c|c|c|c|c|c|}
    \hline
      AGLRFE  & ALPM& CFM & MRFFAM & $F_{max}$ & MAE \\  
    \hline
     $\times$& \checkmark & \checkmark & \checkmark& 0.906 & 0.035 \\
    \checkmark  & $\times$ & \checkmark & \checkmark&  0.907 & 0.033 \\
    \checkmark  & \checkmark & $\times$& \checkmark&  0.908 & 0.032\\
    \checkmark  & \checkmark & \checkmark& $\times$&  0.906 & 0.032\\
    \checkmark &  \checkmark & \checkmark& \checkmark& \textbf{0.917} & \textbf{0.029}\\
    \hline
  \end{tabular}
  \caption{Influence of individual components in SODAWideNet++.}
  \label{tab:components}
 \end{table}

\subsection{Difference from the previous loss pipelines}

 Prior works only focused on supervising the foreground (saliency) maps, whereas we supervised both the foreground (fg) and background (bg) maps. As seen from Table \mbox{\ref{tab:bglosssupervision}}, background supervision slightly improves performance over foreground only supervision.

 \begin{table}[h!]
\centering
  \begin{tabular}{|c|c|c|}
    \hline
      AGLRFE  & $F_{max}$ & MAE \\  
    \hline
     fg + bg & 0.917 & 0.029 \\
     fg & 0.912 & 0.031 \\
    \hline
  \end{tabular}
  \caption{Influence of background (bg) supervision on SODAWideNet++.}
  \label{tab:bglosssupervision}
 \end{table}
\vspace{-12mm}

\section{Conclusion}

In conclusion, our SODAWideNet++ framework integrates the strengths of vision transformers and convolutional networks through the novel AGLRFE module. By using dilated convolutions paired with self-attention mechanisms, our model combines the inductive biases of convolutions and the dynamic, input-specific capabilities of attention mechanisms. This combination identifies salient objects across varied scenes and conditions. We used binarized annotations from the COCO dataset to train the model instead of traditional ImageNet pre-training. This tailored approach aligns directly with the nuances of SOD tasks, resulting in a model with competitive performance across multiple datasets. Our results demonstrate the effectiveness of our proposed pre-training approach and model design choices, where we achieve competitive performance against state-of-the-art models such as RMF \cite{rmformer} while only containing 35\% of trainable parameters.

%
%
%

\bibliographystyle{splncs04}
\bibliography{revision_paper}

\end{document}